\documentclass{jdmdh}
\usepackage{array}
\usepackage{pgfplots}

\title{Pattern Recognition in Narrative: Tracking Emotional Expression in Context}

\author[1]{Fionn Murtagh}
\author[2]{Adam Ganz}
\affil[1]{University of Derby, UK, and Goldsmiths University of London, UK}
\affil[2]{Royal Holloway, University of London, UK}
\corrauthor{Fionn Murtagh}{fmurtagh@acm.org}

\begin{document}

\maketitle

\abstract{Using geometric data analysis, our objective is the analysis of
narrative, with narrative of emotion being the focus in this work.   
The following two principles for analysis of emotion inform our work. 
Firstly, emotion is revealed not as a quality in its own right but rather 
through interaction. We study the 2-way relationship of Ilsa and Rick 
in the movie Casablanca, and the 3-way relationship of Emma, Charles and
Rodolphe in the novel {\em Madame Bovary}.  
Secondly, emotion, that is expression of states of mind of subjects, is 
formed and evolves within the narrative that expresses external events and 
(personal, social, physical) context.  
In addition to the analysis methodology
with key aspects that are innovative, the input data used is crucial.  
We use, firstly, dialogue, and secondly, broad and general description that 
incorporates dialogue.  In a follow-on study, we apply our unsupervised 
narrative mapping to data streams with very low emotional expression.  
We map the narrative of Twitter streams.  Thus we demonstrate map analysis 
of general narratives.}


\section{Introduction}

We establish an approach for mapping out narrative, with a primary focus
on emotional attachment. Two extensively studied testbeds are used.   Our 
methodology does not seek to isolate the qualities of interest.  Rather,
we map these qualities (i) through their interaction, and (ii) in context.

The new interdisciplinarity that we pursue here is between computational 
science and psychoanalysis.  Longer term our aim is data-driven generative
modelling.   This is a different line of work relative to computational 
and psychological interdisciplinarity (\citet{reisenzein}), largely 
using behavioural modelling.  There are strong parallels here, in our view, 
with, respectively, unsupervised and supervised pattern recognition.  Our 
approach is unsupervised, or data driven.

An interesting psychoanalytic view of the link between emotion and
interaction is in \citet{buchholz}:    
``Emotions are a multifaceted object of interest. They should not be 
considered as
producers of interaction. It might be more correct to reverse the order 
of causation: social
interaction is the medium with the power to evoke and regulate human 
emotions.''

In brief, we carry out a metric embedding of the semantics of the 
discourse, and determine subplots of the narrative.  From the metric 
embedding of the semantics we map the trajectory of emotional 
qualities, and emotional interactions of the characters.  Also from 
the metric embedding of the semantics, we structure the continuity and 
change in the narrative using a hierarchy. 

In section \ref{litreview}, we present a short review of 
various strands of work in this area, that have lessons to be drawn. 
We indicate where, and why, our work differs.

Section \ref{unsuppattrec} deals with our methodology of unsupervised
pattern recognition.  

Section \ref{emotint} relates to subplot analysis in the dialogue of
Ilsa and Rick in the Casablanca movie.  We ask: how is the overall
dialogue of Ilsa and Rick to be understood in terms of subplots?
Such subplots are found by us in decreasing order of importance or 
salience.  They are determined in terms of factors, i.e.\ principal components or 
latent semantics.  
In section \ref{casasent} we analyze emotion through emotional interaction.
From the data, we find the major determinants of emotional interaction.  
This work is based on dialogue from the Casablanca movie. Since this has been 
very extensively analyzed, it provides us with an evaluation testbed.  

Section \ref{mmebov} is a study of the narrative and evolution 
of emotion from part of Flaubert's 19th century novel, {\em Madame Bovary}.  
Since we use text from this novel, with description 
as well as dialogue, this case study shows how well our methodology can handle
general textual data.  

In section \ref{twitter} we show the applicability of our unsupervised
(or data-driven) methodology to very general narratives.  
We apply the principles of the methodology developed
to Twitter streams.  We map out the semantics through the Correspondence Analysis
mapping of input frequency of occurrence data into the factor space.  Next we 
consider sub-narratives from (statistically significant) segmentation of the 
Twitter stream.  We locate, and then contrast, emotive words in the 
semantically-embedded structuring of the overall narrative.  

\section{Review of Determining and Tracking Emotion in Various 
Application Domains}
\label{litreview}

We review in this section a number of approaches to determining and 
tracking emotional context in text and/or related data.  

In subsection \ref{reviewwlv}, lexical analysis based machine learning 
is used on social media data in order to label  such data in terms 
of emotion or mood.  
In subsection \ref{revieweastsf} lexical analysis is used on novels in 
order to subtype them, and otherwise label them, in terms of emotional
descriptors.  The extrapolation of such an approach is proposed 
for characterizing the
development of indications towards anti-social behaviour.  
Subsection \ref{reviewtee} uses a hierarchy of mood states, that is 
manually created, and this is applied in a dynamic, evolving context.  

\subsection{Supervised Lexical Analysis for Sentiment and Mood Analysis}
\label{reviewwlv}

In \citet{thelwal}, positive or negative sentiment towards consumer
products is the focus.  

Two approaches are used -- ``mood setting'' which characterizes social 
web texts on scales
of $+1$ to $+5$, and $-1$ to $-5$, for selected sentiment words, with 
some context taken into account, such as questions, negations, exclamation
marks, emoticons, and the use of the adverb ``very''.  ``Lexicon extension''
(also referred to on page 3 of \citet{thelwal} as lexical extension) seeks
other additional words that are to a greater or lesser extent characteristic
of alternative moods.

In \citet{thelwal2}, rather than positive and negative scales of sentiment, 
``valence'' and ``arousal'' are studied.  
In \citet{genereux}, the terms ``mood'', ``affective state'' and ``emotion'' 
are identical.  Schwartz's typology of motivational values is used, and 
these are further taken on two dimensions, positive/negative and 
passive/active.  This study associated moral value terms
with mood.  Mood was defined by degree of the motivational values.   
Marked corpora were used.  

Relative to the work described, which we can characterize as 
lexical and supervised, our work is unsupervised and is data 
driven, in particular insofar as uncontrolled term sets are 
determined through our pipeline processing and analysis.  A further 
and crucial point of note is how the work described takes emotion 
as, very largely, mood, with a view towards prediction of a 
user's behaviour.  Such behaviour can be purchase of a commodity,
or an anti-social act.  

\subsection{Emotion Tracking through Lexical Analysis}
\label{revieweastsf}

In the emotion tracking project, \citet{sutinen}, 
the approach taken is to determine
terms and phrases that carry sentiment semantics, and then label 
them in degree of positivity or negativity.  Examples of application
areas are marketing and conflict prediction.  The latter entails
the search for ``drastic changes in emotions towards a certain 
topic'', \citet{sutinen}.  (Multiple murder in a school shooting is 
at issue here, and in subsection \ref{reviewwlv} above, street rioting in the
UK was at issue.)

It is noted in \citet{kakkonen} how sentiment analysis, as a new 
area of text analysis, has come to the fore in recent years.   Emotion 
or affective content in novels is the focus, together with visualization.
This work studies Gothic novels, that are ``rich in dark and gloomy 
topics'', hence negative emotions, and with sub-genres of terror and 
horror. A pre-established class hierarchy of nouns, verbs and adjectives
is used.  As an example, the ``cruelty'' class was associated with such words
as ``cruel'', ``pitilessness'', ``unkind''.   Sentiment-bearing words 
are scored, and these scores can be used to distinguish between the 
sub-genres noted, of terror and horror. 

Where our work differs from \citet{kakkonen} is that we seek overall semantic 
characterization from data in an unsupervised, rather than supervised,
way.  

\subsection{Narrative of Emotion together with Narrative of Actions}
\label{reviewtee}

In this subsection, we review work focused on storyboarding or 
dramatization, together with plot unit summarization.   

In addition to the storyline defined in \citet{pizzi07} as a sequence of
actions (designed to address a task that the story defines), referred to 
as the plot, this work seeks to use ``character'' (\citet{pizzi07}, section 
1) also.  In the plot, there is generation of action sequences.  With 
character, the plot's actions are to be driven by intention and longer
term motivations.   

In an analysis of three chapters from Flaubert's 19th century novel, 
{\em Madame Bovary}, \citet{cavazza09}, a top level polarity of feelings 
(states of mind) into ``duty'' and ``pleasure'' is used; and a set of 
five ``emotional input'' feelings, being combinations of ``valence'' 
and ``arousal'', that are determined in real time 
from the voice of a user interactively playing the role of Emma Bovary's 
lover, Rodolphe. These emotional input feelings are: NegativeActive, 
NegativePassive, Neutral, PositiveActive and PositivePassive.  

In discussing other work, \citet{cavazza09} (in their section 6)
note the objective being 
simulation and training, conferring believability on the virtual actors, 
while not at issue are narrative as such, or aesthetic aspects.  It is 
acknowledged (in their section 7) that the emotional categories, for 
both state of feelings and expectations, are limited in number and 
genre-specific.   

The modelling approach, originating in \citet{schank},
has led in the work of \citet{lehnert} to determining plot units from
actors and actions, through use of ``affect states'' that characterize
them, and weighted links that provide causal and cross-character values.
Such a supervised approach is library-based (i.e.\ library of primitive plot units), 
with a taxonomy of affect states.  \citet{goyal} take such modelling 
considerably further, and include discussion of how to recognize affect states.
In \citet{goyal}, 34 of Aesop's fables are used to exemplify this.  Such a
supervised modelling approach to the creation of plot unit structures is relevant in
well-defined and well demarcated contexts.

\subsection{Dynamical Systems Modelling of Emotional States}

\citet{rinaldi}
model the Walt Disney film, ``Beauty and the Beast'', using state 
variables that represent feelings, and appeal functions.   
Emotions of partners in a relationship are modelled using ordinary 
differential equations.  The dynamical system that ensues is then 
explored in terms of its system properties (such as chaotic behaviour).  
Parameter values are provided in analogy with the film content.  

\citet{khrennikov2,khrennikov} has developed and explored dynamical systems
in p-adic (p prime) number systems.  
Cognitive process dynamics, represented as paths and trajectories 
in trees, imply that (see \citet{khrennikov} p.\ 138)  a ``pathway
contributes to distinct psychological functions'', and furthermore 
that ``pathways going through reasoning-centres can go through some 
emotional centre.  Thus reason participates in the creation of emotions
and vice versa.''   
Such dynamical systems work is not data-driven as in our approach.  

\section{Emotion Analysis through Unsupervised Pattern Recognition 
in Textual Data Usage}
\label{unsuppattrec}

In this section, the principles of our methodology are discussed, together
with various aspects relating to implementation.  

For \citet{mckee}, text is the ``sensory surface'' of the underlying
semantics.  This is also what \citet{benz79a,benz79b,benz81} 
referred to as ``letting the data speak'', and
``the model must follow the data and not the other way around''. 
Our approach is one of (i) data encoding, (ii) measurement,
including especially qualitative information through proxied 
expression of such information, (iii) 
pattern and trend determination and display, and thereafter 
(iv) model selection from the data.  

\subsection{Euclidean Geometry for Mapping the Semantics, 
Tree Topology for Chronology of the Semantics}
\label{method}

Our starting point is the mapping of semantics into a visually-based 
representational space.  This is accomplished by considering text (or 
other multimedia data) units crossed by the entities that comprise them
and that serve to define them -- words, in the case of text.  Thus each text 
unit, e.g.\ a chapter of a novel, or scene of a filmscript, or one or 
some consecutive set of Twitter tweets, and so on, is contextually 
described by the frequencies of occurrence (including absence) of words,
in the word set used.  Similarly, each word has a vector of the 
text units in which it appears.  Context, therefore, is part and parcel 
of our data from the start.  

The choice of words in the word list
includes -- without loss of generality -- setting all upper case to lower
case and excluding punctuation. 

The dual spaces of text units and words are then mapped mathematically
(and analytically using the eigenvalues and eigenvectors of these dual 
spaces, defined from the positive semidefinite matrix of their 
relationships) into a Euclidean space, termed a factor space (or principal
component space, or latent semantic space). Planar projections of the
factor spaces provide visualizations of the text units and words of the
dual spaces.  In \citet{harcourt} this is described as map analysis of 
qualitative observed data.  
These planar projections are based on decreasing informativeness of the 
factors, that are also referred to as space dimensions, or axes.  
Mathematically, the informativeness of the factors is defined by the 
moments of inertia of the clouds of points that are analysed.  These 
are the dual clouds (mutually defining one another), the cloud of textual 
units employed, and the cloud of characterizing words.  Through contribution
to the inertia of a factor, we can read off the most contributing text 
units, and the most contributing words.  

On the basis of the representation of the field of study, analysis of 
change over time, including continuity and change, next avails of a 
hierarchical mapping.  We can require that the hierarchy be based on 
the chronology, or given sequence, of text units (e.g.\ dialogue elements, 
descriptive text segments). 
The algorithmic requirements to realise this (the pairwise agglomerative
criterion that ensures a well-defined hierarchy, viz.\ with a monotonic 
sequence of agglomeration values: the complete linkage criterion) 
are described in \citet{murtagh2009}.  Typically the full dimensional 
field representation (i.e.\ factor space) is input to the hierarchy 
building algorithm.  

For clustering, we are taking multiple scales into account through 
use of agglomerative hierarchical clustering.  Secondly, rather than 
a typological analysis, what we want is to base the clustering on the 
chronology, or given sequence, of our text units.  A further benefit of 
this approach is that, if required, we can statistically assess whether 
or not to carry out an agglomeration.  This is based on a permutation 
test, in order to assess the statistical significance of a merger.  
This methodology is described in \citet{becuebertaut}.  This furnishes
a succession of clusters in our sequentially ordered set of text units.
The succession of clusters is a segmentation of our time-line or overall
narrative (expressed by the text units).  

Sequence-constrained hierarchical clustering will be shown, below, to 
represent change in narrative direction.  As noted, it takes scale into 
account.  Relative to our narrative, this is scale of continuity and change.
Using a basis in statistical significance testing, we can segment the 
narrative.
 
In summary, (i) our initial data is endowed with the $\chi^2$ metric (based 
on frequency counts, including zeros) in order
to account for all semantic information; (ii) the dual (text, word) spaces are mapped
into the factor space, and are endowed with the Euclidean metric (which can be 
said to be the natural, visual distance); (iii) 
the dual spaces, now in their Euclidean geometry, are mapped into an ultrametric 
topology.  Ultrametric topology expresses hierarchy, or a tree structure.  


\subsection{Analysis of Emotion: The Importance of Tool or Function Words}
\label{toolwords}

In general we can distinguish between (i) tool words, form words, 
function words, empty words, grammar or grammatical words
(examples: articles, prepositions, conjunctions, pronouns,
and so on); and (ii) content words, full words.  In \citet{murtagh05}, we 
cite many studies, using many languages, to support the following.  Tool words 
are of importance for style analysis, and full words for content analysis.  
However style itself can be useful for describing context and relative properties.  
In particular ways, we can approach the analysis of content through tool words.
Non-tool full words can be very specific,
and therefore not at all helpful for comparative exploration of texts or
text fragments.  

Content words, with lemmatization and stemming, are important in supporting 
retrieval.  
Other forms of style and content can be relevant too, such as sentence length, 
patterns of punctuation usage, word length, use of the passive tense, etc. 
For us we make the following choice, given our aim in this work.  The ``texture'' of the 
text is very important.  \citet{pennebaker},
and see also \citet{moyer}, analyzed ``junk'' words (articles, pronouns, prepositions)
and showed how they typified gender, and personal relationships (in the heading of 
\citet{moyer}, ``word usage predicts romantic attraction'').   

Thus we have a particular interest in tool or function words.  In this context, there 
is no benefit to us in lemmatization and stemming.  Note also that 
the very limited number of words used in dialogue precludes this. 

In section \ref{twitter}, analysis of Twitter data, 
we employ a stoplist to remove function words. 
This is because there is far less concern for emotion-laden words. 

\section{An Analysis of Emotional Interaction in the Movie ``Casablanca''}
\label{emotint}

In the movie Casablanca, to study emotion we look here specifically at the 
two main protagonists, or characters, Ilsa and Rick.   

\subsection{Filmscript Text Data Used}

From the 77 scenes in Casablanca, we took the scenes where there was 
presence of both Ilsa and Rick.  These were scene numbers 22, 26, 28, 
30, 31, 43, 58, 59, 70, 75 and 77.  We selected just the dialogue
of the two, Ilsa and Rick, and in addition required that this dialogue 
be alternating. (There were 
some cases within a scene involving Ilsa and Rick where the dialogue 
of one of these involved other characters: such cases were not included,
given our objective of focusing on direct Ilsa and Rick conversations.)   
Thus, dialogue elements were alternating between the two, Ilsa and Rick, 
with just a few instances of repeated utterances of the same person (Ilsa
or Rick).  

In all there were 150 dialogue utterances.  The number 
of unique words in these utterances was 528, with: upper case set to lower case, 
punctuation set to blank, words constituted of 
$\geq 1$ contiguous letters.  
The number of these words that were used, including the speaker name for the 
utterance, was 2435.   
Next we required, to support the contrasting of utterances, 
that a word should be used a sufficient number of times.  We required 
that each word be used in at least two different utterances, and that each 
word be used at least twice overall.  This gave the total number of words
retained as 261.  

The data input is a {\tt csv} file.  We have made it available to the reader 
at www.narrativization.com. The first few dialogue utterances 
and the very final one 
are as follows, where we have Sequence Number (of the utterance), Name (Ilsa
or Rick), and the Expression -- the utterance.  The dashes are as in the original
script.  

\medskip

{\small
\noindent
1, Rick: ``-- Hello, Ilsa.''

\noindent
2, Ilsa: ``Hello, Rick.''

\noindent
3, Ilsa: ``-- This is Mr. Laszlo.''

\noindent
4, Rick: ``How do you do?''

\noindent
5, Ilsa: ``I wasn't sure you were the same. Let's see, the last time we met --''

\noindent
6, Rick: ``-- It was La Belle Aurore.''

\noindent
7, Ilsa: ``How nice. You remembered. But of course, that was the 
day the Germans marched into Paris.''

\noindent
8, Rick: ``Not an easy day to forget.''

\noindent
9, Ilsa: ``No.''

\noindent
10, Rick: ``I remember every detail. The Germans wore gray, you 
   wore blue.''

\noindent
11, Ilsa: ``Yes. I put that dress away. When the Germans march 
   out, I'll wear it again.''

\noindent
...

\noindent
150, Rick: ``You better hurry, or you'll miss that plane.''
}
%
%
%
%
%
%
%
%
%
%
%
%


\subsection{Semantic Mapping of the 150 Dialogue Utterances}
\label{semutt}

\begin{figure*}
\centering
\includegraphics[width=9cm]{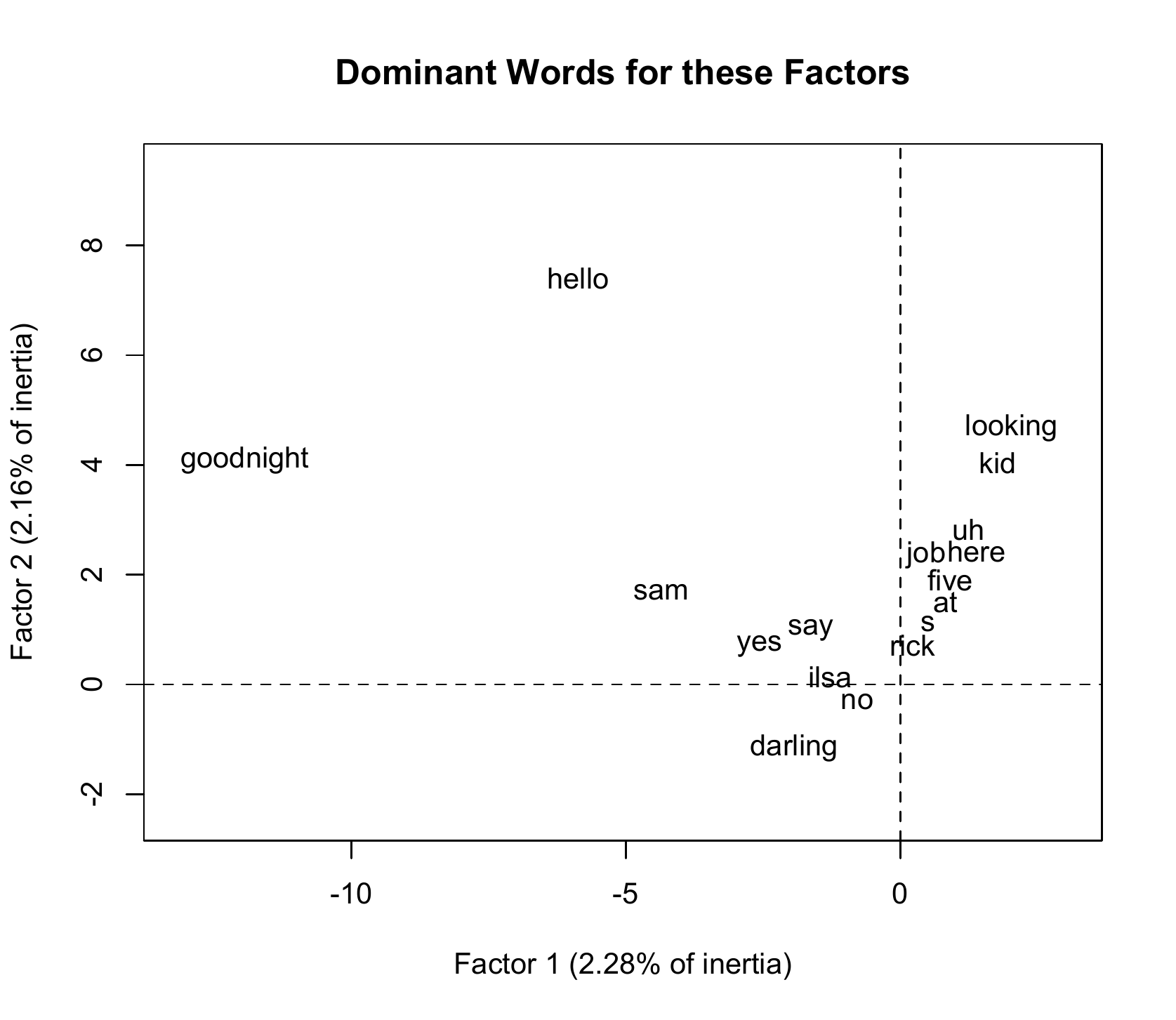}
\caption{Principal plane of factors 1 and 2, resulting from the 
Correspondence Analysis, with the most contributing words (out of the 261 words)
displayed. ``s'' is ``is'',
resulting from, e.g., ``here's''. The words ``Ilsa'' and ``Rick'', in this display,
are words used in the dialogue itself.}
\label{DomWordsF12}
\end{figure*}

\begin{figure*}
\centering
\includegraphics[width=9cm]{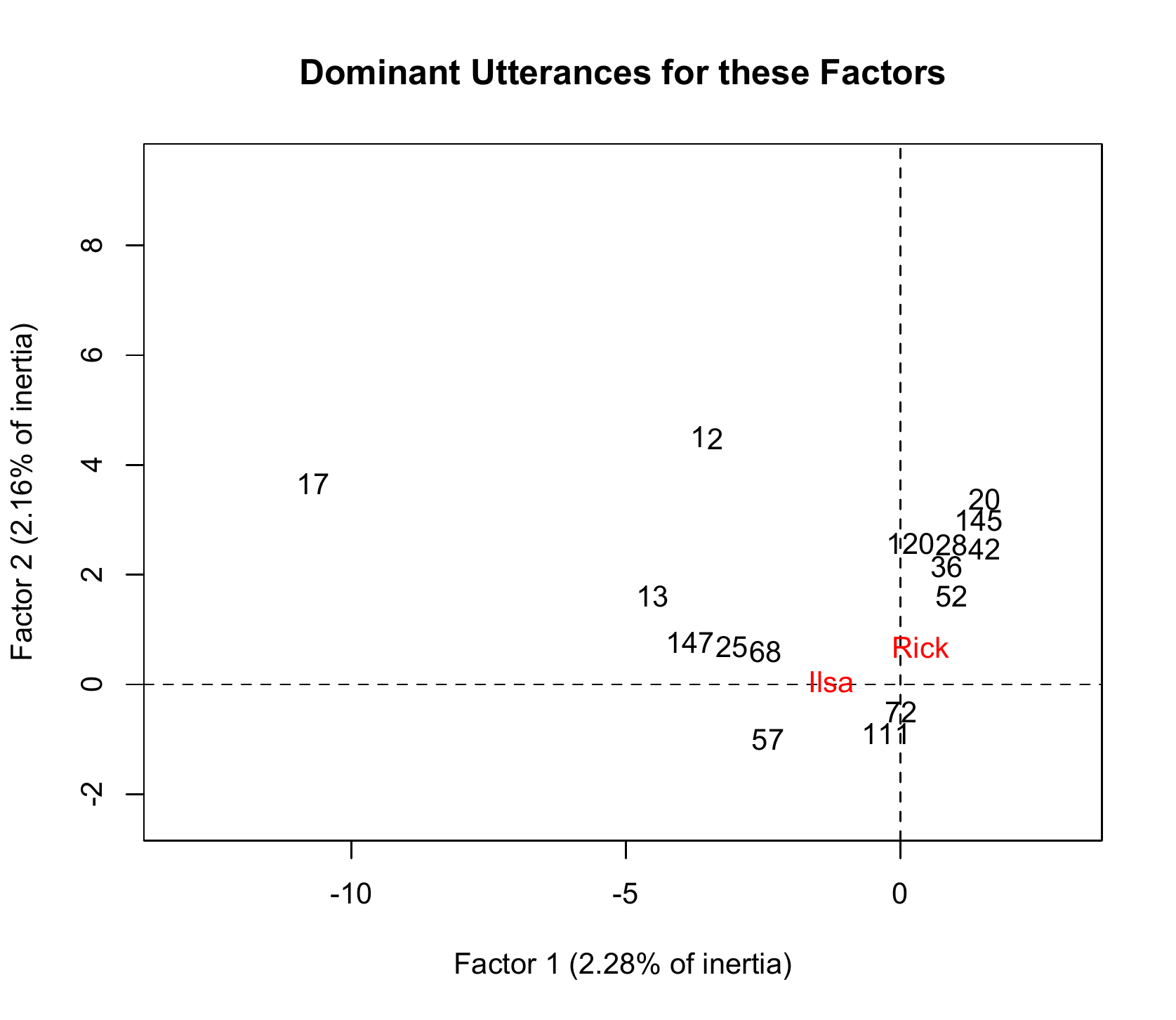}
\caption{Principal plane of factors 1 and 2, resulting from the 
Correspondence Analysis, with the most contributing utterances (out of 150)
displayed. The names ``Ilsa'' and ``Rick'', in this display, are supplementary 
attributes and are the speaker name, that is separate from the use of the person's
name in a dialogue utterance.}
\label{DomUttF12}
\end{figure*}

Our first analysis is frequency of occurrence cross-tabulations of the 150 utterances 
with the 261 word set.  Figures \ref{DomWordsF12} and \ref{DomUttF12} are separately 
displayed simply to avoid overcrowding. 
The factors (axes, components) are the principal axes of inertia of the 
cloud of weighted utterance points, and this overall cloud inertia is identical to 
the inertia of the cloud of weighted word points.  
The projections on this best fitting, principal factor, plane 
represent just 2.28\% (first factor) 
plus 2.16\% (second factor) of the inertia of the cloud of 150 utterances, or 
identically the inertia of the cloud of 261 words.  This is typical for the sparse
utterance/word data that is analyzed.  

To focus our attention on the most influential utterances and words, the contribution 
to inertia of the factors is checked out.  A threshold of 3 times the mean contribution 
was used here.  This was a trade-off between decreasing inertia contributions, versus 
the need to restrict attention to a limited number of utterances and of words.  
Figures \ref{DomWordsF12} and \ref{DomUttF12} show words and utterances that have 
inertia contributions that are greater than 3 times the mean (word or utterance, 
respectively) contribution.  In Figure \ref{DomUttF12}, utterances 17, 1, 2, 13, 68, 
25, 147, 57 contributed strongly to the inertia and are on the negative side of the
first factor.  There were no single 
utterances contributing strongly to the positive side of 
the first factor.  For the second factor, utterances 20, 145, 28, 42, 36, 52, and 
also 1, 2, 13, contributed strongly to the positive side of this factor.  Contributing
strongly to the negative side of factor 2 were utterances 72, 111.  

While Figure \ref{DomWordsF12} has the names ``Ilsa'' and ``Rick'' in the word set,
i.e.\ among the 261 words, in Figure \ref{DomUttF12} we do the following.
We use these names, i.e.\ the speaker of each of the utterances (coded for each 
utterance as 1 if the speaker, and as 0 if not the speaker), as supplementary 
attributes.  That is, they 
are projected into the factor space subsequent to the determination of the factor 
space.  They are passive attributes.  It is interesting to note how these
supplementary words have very close projections in the principal factor plane 
(Figure \ref{DomUttF12}), similar to when they are used as words in the 
dialogue itself (Figure \ref{DomWordsF12}).  

The utterances in Figure \ref{DomUttF12}, that are the most important for this principal 
factor plane, are interesting as shown in the following list.

{\small
\begin{itemize}
\item Positive factor 1, positive factor 2 (nostalgic and endearing).
\begin{itemize}
\item 
20, Rick: ``Here's looking at you, kid.''  \\
145, Rick: ``Here's looking at you, kid.'' \\
36, Rick, ``You said it! Here's looking at you, kid.'' \\
120, Rick, ``All right, I will. Here's looking at you, kid.''
\item 
28, Rick: ``Uh huh.''  \\
42, Rick: ``Looking for a job.''
\item 
52, Rick: ``-- The train for Marseilles leaves at five o'clock. I'll pick you up 
at your hotel at four-thirty.''
\end{itemize}
\item Negative factor 2, positive factor 1 (discursive and debating).
\begin{itemize}
\item 
72, Ilsa, ``Last night I saw what has happened to you. The Rick I knew in Paris, I 
could tell him. He'd understand. But the one who looked at me with such hatred... 
well, I'll be leaving Casablanca soon and we'll never see each other again. We knew 
very little about each other when we were in love in Paris. If we leave it that way, 
maybe we'll remember those days and not Casablanca, not last night.'' \\
111, Ilsa, ``Oh, it wasn't my secret, Richard. Victor wanted it that way. Not even our closest friends knew about our marriage. That was his way of protecting me. I knew so much about his work, and if the Gestapo found out I was his wife it would be dangerous for me and for those working with me.''
\item 
57, Ilsa, ``Oh, darling!''
\end{itemize}
\item Negative factor 2 (positive exchange, tending to being formal).
\begin{itemize}
\item 
68, Ilsa, ``Yes.''  \\
25, Ilsa, ``Yes?''  \\
147, Ilsa, ``Yes.''
\item 
13, Ilsa, ``Say goodnight to Sam for me.''
\item
1, Rick, ``-- Hello, Ilsa.'' \\
2, Ilsa, ``Hello, Rick.'' \\
17, Ilsa, ``Goodnight.''
\end{itemize}
\end{itemize}
}

Our semantic mapping has indicated the importance of 
the famous catch-phrase of the Casablanca movie 
script, ``Here's looking at you, kid''.  We also 
note the importance of the one-word dialogue utterances, ``Yes''.  We repeat a 
point made earlier: with such short utterances, 
we are truly studying the ``texture'' of the dialogue.  

We note how important Sam, the piano player, is here in mediating the relationship
of Ilsa and Rick (cf.\ upper left of the principal factor plane).  The upper left
quadrant is a somewhat formal and friendly space, referenced to the background context.
In that sense, the Ilsa/Rick relationship is very much mediated by the ambient context. 
The upper right 
quadrant is, we could say informally, Rick's space.  It is here where we find his 
celebrated, and repeated, phrase ``Here's looking at you, kid.''  The lower left
quadrant is, again informally, Ilsa's space, with a somewhat more amorous tendency. 

We would characterize the factors as follows.  Factor 1, negative, is dominated
by an openly, positive, Ilsa.  Factor 1, positive, is dominated by an endearing Rick.
Factor 2, negative, is dominated by debating.  Factor 2, positive, is dominated
by comfortable exchange. 

Were we to have a straightforwardly maturing relationship between Ilsa and
Rick, then we might find the chronology of utterances in some simple pattern in 
the principal factor plane.  We do not find such a pattern, because the 
relationship between these characters evolves through the film.  The narrative and
the subplots of the narrative are what we look at next.  

\subsection{Semantic Analysis through Euclidean, Factor Space Embedding: 
Application to Determining the Most Salient Subplots}
\label{subplot}

In order to examine the subplots that are interwoven into the overall narrative, 
we proceed as follows.  The utterances, which are relevant for the emotional 
content as we have shown in the previous subsection, are not always 
suitable directly for 
studying the evolution of the narrative.  (After all, there are a few utterances 
that just have ``Yes'' and ``No''.  But in an emotional context, clearly they have their
role.) 

So, to study the evolution of the narrative, we use scenes.  The scenes, and the
dialogue utterances that come from them, are listed in Table \ref{tabscutt}.  From 
the input data point of view, we aggregate the utterance data to provide the 
scene data.  Therefore, we have up to now 
a $150$ utterances $\times$ $261$ words table with 
0 values whenever the word was not used in the utterance, and a value greater than 
0 for the number of times the word was used in the utterance.  (The largest such
frequency of occurrence is 7, and this is the case for ``you'' in utterance 95, 
and for ``I'' in utterance 113.)  We aggregate rows 1 to 17 to form the new row
of the scenes matrix, and so on (cf.\ Table \ref{tabscutt}).  This gives us a 
frequency of occurrence, scenes $\times$ words, matrix of dimensions $11 \times 261$.  
This $11 \times 261$ matrix has 910 non-zero values and therefore has 31.7\% occupancy. 
Its maximum value is 43 (``you'' is used 43 times in one scene).  

\begin{table}
\begin{center}
\begin{tabular}{|l|rr|} \hline
Scene &  \multicolumn{2}{c|}{Utterance}   \\ 
      & First in scene & Last in scene \\ \hline
22    &   1 & 17 \\
26    &  18 & 20 \\
28    &  21 & 31 \\
30    &  32 & 34 \\
31    &  35 & 60 \\
43    &  61 & 77 \\ 
58    &  78 & 107 \\
59    & 108 & 121 \\
70    & 122 & 128 \\
75    & 129 & 145 \\
77    &  146 & 150 \\ \hline
\end{tabular}
\end{center}
\caption{The scene numbers (from the 77 scenes in the Casablanca script) with
the first and last of the utterances that are from that scene.  The utterances
are (mostly consecutively) by either Ilsa or Rick, thus being dialogue expressions between
these two characters.}
\label{tabscutt}
\end{table}

The factors, as the principal components, or latent variables, are the 
bearers of the information that underlies our data. They are given in order of
decreasing importance.  Therefore we will determine subplots in the overall 
narrative by looking at the factors, in succession.  Since we can do that by 
displaying sets of planar projections, and also to economize on our discussion,
we will discuss our findings by pairs of successive factors.  
We summarize our findings as follows, for the 11 scenes (and 261 words).  

\noindent
{\bf Factors 1 and 2:}
The most important scenes (in terms of contribution 
to the inertia defining the principal axes of inertia, or factors) 
are scene 22 and scene 31.   Scene 22
is where Ilsa and Rick meet again in Casablanca.  Sam the piano-player and singer 
is present.  At one point we have: 
``ILSA: Play it, Sam. Play `As Time Goes By.' ''.  Later there is this 
exchange: ``RICK: Hello, Ilsa. ILSA: Hello, Rick.''  There is 
discussion of Germans in Paris in the past. The scene ends with Ilsa and
Rick saying ``Goodnight'' to one another.  In scene 31, there is a 
flashback to the past in Paris, with Sam there too, and ending with:
``ILSA: Kiss me. Kiss me as if it were the last time.''.  \\
Therefore the positive quadrant for both factors is 
introductory and scene setting, based around Sam.  Factor 2, negative
half axis, refers to the past, in scene 31, 
and the positive half axis
refers to the present, in particular scene 22.  

\noindent
{\bf Factors 3 and 4:}
In the plane of factors 3 and 4, the preponderant scenes are
as follows. Scenes 58 and 59, 
relate to Ilsa's husband Victor Laszlo (``he'') getting transit 
papers to flee from Casablanca.  Scenes
76 and 77 are at the airport where Rick sees off on the plane both
Ilsa and Victor.  \\
Whereas the plane of factors 1 and 2 relates to Ilsa and 
Rick meeting up again, the plane of factors 3 and 4 relates to their separation 
from one another.  

\noindent
{\bf Factors 5 and 6:}
In the plane of factors 5 and 6, 
scene 28 features, and this scene has Rick and Ilsa 
in a flashback to Paris, Rick in love with Ilsa stating that Victor 
Laszlo is dead, as she thought then.   This plane also has as a 
preponderant scene, scene 77 which 
has Rick shooting dead Major Strasser, in self defence, and in order 
to allow Ilsa to escape from Casablanca with Victor.  \\
If the planes of factors 1 and 2, and of factors 3 and 4, are 
respectively the beginning and the end of Ilsa's and Rick's renewed relationship, 
then the plane of factors 5 and 6 
is a succinct, pithy summary, through the scenes displayed, of the 
overall narrative of this renewed relationship.  

\noindent
{\bf Factors 7 and 8:}
In the plane of factors 7 and 8, preponderant 
scene 28 is a Paris 
flashback, with a very short dialogue where Rick is in 
discovery mode about Ilsa (with ``Who are you really?'', and other 
questions to her).  Then scene 43 is the
market scene that we analyzed in \citet{murtagh2009}.  
That latter scene is the time when Ilsa announces that she will 
leave Casablanca,
and that Victor Laszlo has been her husband all along.  Again therefore
Rick is put (unwillingly) into discovery mode about Ilsa. 

\noindent
{\bf Factors 9 and 10:}
The plane of factors 9 and 10 counterposes scene 30, which 
in Paris points to Rick having to flee for safety, and Ilsa's 
concern for Rick's safety; with 
scene 70, where Rick indicates that he is helping both 
Victor and Ilsa flee for safety, and again Ilsa's concern for 
Victor is uppermost.  

In summary we have the following.

\begin{enumerate}
\item The plane of factors 1 and 2: 
the beginning of the renewed meeting up of Ilsa and Rick in 
Casablanca, with the Paris background to this.  
\item The plane of factors 3 and 4:
the end of the renewed meeting up 
of Ilsa and Rick in Casablanca. 
\item The plane of factors 5 and 6:
a focus on what Rick did in the 
three-way relationship between Ilsa, Rick and Victor.  
\item The plane of factors 7 and 8:
subplot of Rick in discovery mode 
about Ilsa, actively back in Paris, and with Ilsa in the more 
active, revealing role, now in Casablanca.  In the latter 
revelation, Victor enters decisively into their -- Ilsa's, Rick's 
-- relationship.    
\item The plane of factors 9 and 10:
subplot of Ilsa's concern for Rick, 
Ilsa's concern for Victor, and Rick's crucial support of Victor.  
\end{enumerate}

Our semantic analysis thus far has been firstly to determine emotional 
interaction between Ilsa and Rick, which was done through analysis of 
dialogue utterances (subsection \ref{semutt}).  Then (this subsection
\ref{subplot}) we determined subplots in the overall narrative.  

\section{Chronology of Emotional Interaction}
\label{casasent}

The following two principles for analysis of emotion inform our work. 

\begin{itemize}

\item Emotion is revealed not as a quality in its own right but rather 
through interaction. Hence our interest in the direct interaction 
of the two characters, Ilsa and Rick. 
\item Emotion, that is expression of states of mind of subjects, is 
formed and evolves against the backdrop of external events and (personal,
social, physical) context.  Hence we traced out subplots (in subsection 
\ref{subplot}) because they define this context. 
\end{itemize}

We now use the full factor, latent semantic embedding space.  Hence no 
reduction in information is at issue, e.g.\ through a low dimensional semantic 
approximation.  

\subsection{Evolution and Strength of Sentiment}

Our approach to tracking evolution of sentiment is unsupervised, 
i.e.\ data-driven.  This differs from supervised approaches, that 
aim at classifying, in the sense of identifying, words as being of 
some label.  For example, the {\tt classify\_emotion} program of 
\citet{jurka} in the R programming language, assesses text for 
labelling as one or more of these terms: ``anger'', ``disgust'', 
``fear'', ``joy'', ``sadness'', ``surprise''.  A trained naive Bayes 
classifier is used.  Our interest in unsupervised analysis is in 
order to map the potentially very great complexity of what constitutes
emotion.  Ultimately, our interest is not in sentiment as mood but
rather emotion as sentiments that express unconscious
thought processes.  See \citet{langpsych}.  

\begin{figure*}
\centering
\includegraphics[width=9cm]{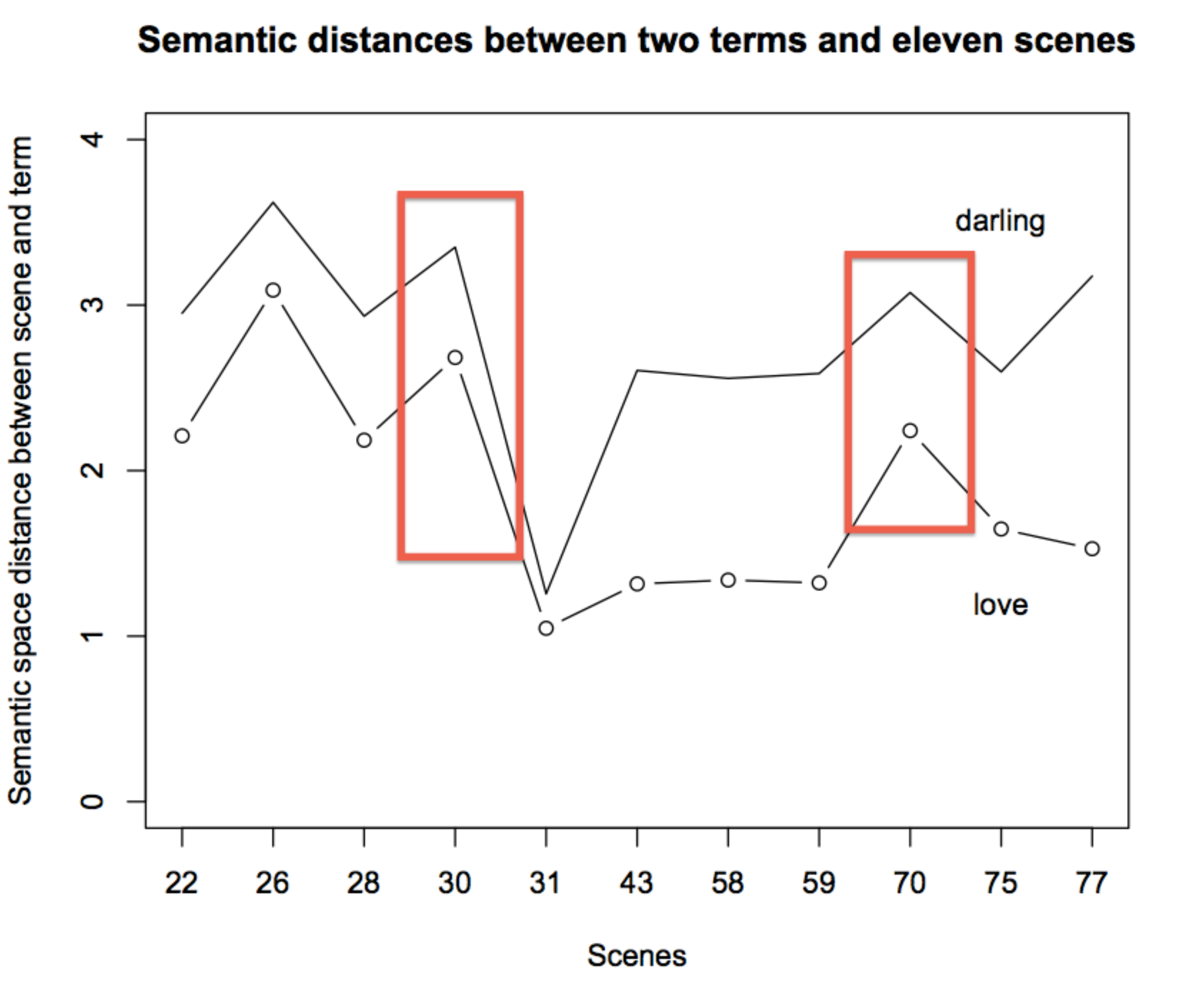}
\caption{In the full dimensionality factor space, based on all 
interrelationships of scenes and words, we determined the distance 
between the word ``darling'' in this space, with each of the 11 scenes
in this space.  We did the same for the word ``love''.  The semantic 
locations of these two words, relative to the semantic locations of 
scenes 30 and 70 are highlighted with boxes.}
\label{figsentiment}
\end{figure*}

Let us use particular words as proxies of strong emotional feeling.  
We recall that dialogue alone is at issue here (no metadata, nor 
any involvement by other characters), in scenes where Ilsa
and Rick were exchanging successive utterances with each other.  
The word ``kiss'' was rare in the dialogue and therefore not retained for 
analysis.  We used the words ``darling'' and ``love''.  Figure 
\ref{figsentiment} indicates that scene 31 is 
highest in terms of emotional attachment, as indicated by both
of the sentiment-bearing words that we used.  The last utterance
recorded for Ilsa in this scene is: 
``Kiss me. Kiss me as if it were the last time.''  In scene 70, Rick 
assures Ilsa that he will help Victor 
Laszlo to escape, and thus there is less of a direct and immediate
emotional attachment between the pair in that scene.  The d\'enouement
in the very last scene points to close emotional attachment, 
proxied by the word ``love'', as Ilsa and Victor fly away.  

We see therefore how easy it is to track sentiment or other 
emotional states, through sentiment-bearing or sentiment-expressing terms.

We additionally tracked, by scene, the terms ``Rick'' and ``Ilsa'' in the dialogue. 
We found these names as used in the dialogue to closely track the swings of 
movement ranging over low values in Figure \ref{figsentiment} implying
strong emotional attachment (e.g.\ in scene 31), and swings away from emotional
attachment (e.g.\ in scene 70).  We also found the differences, by scene, 
to be very small between Rick and Ilsa, again using these names from the dialogue.
We did find differences, with Rick more imposing in scenes 22, 28, less
so in 31, 58 and 70 (note that what we found was every second scene, with the 
exception of the very last), relative to Ilsa who was more imposing in all 
other scenes.  However these differences between Rick and Ilsa were small.  

\subsection{Ultrametric Topology of Narrative}
\label{sectVC}

The narrative of emotion, as expressed by the data we are using here,
can be tracked in a parsimonious way by mapping the full dimensionality
factor space data into a topology that respects the sequential or 
chronological order of the scenes.  We use a sequence-constrained, 
or chronological, agglomerative hierarchical clustering algorithm.
 
In this particular case, we have found it beneficial to 
endow the correlation set, of each scene with all factors, with this
ultrametric topology.  Implementation-wise, this means that we use 
for each scene its set of correlations, i.e.\ cosines squared between the point
and the axis (rather than the set of projections, or
coordinates) in the factor space.  The reason for doing this is to 
focus on relative orientation of the narrative, rather than on the evolving 
relative position of the narrative.  

\begin{figure}
\centering
\includegraphics[width=8cm]{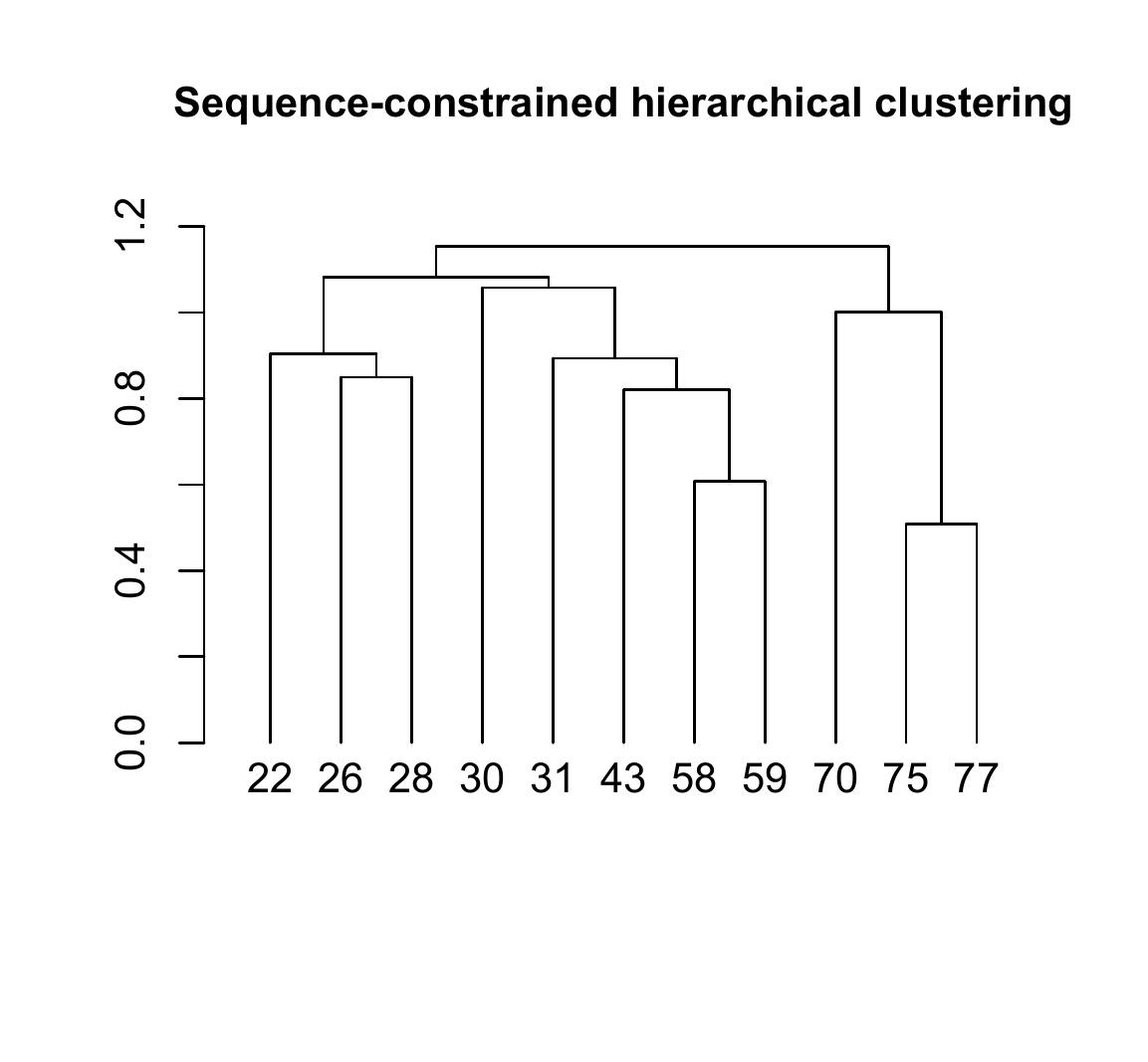}
\caption{Hierarchical clustering, that is sequence constrained, of the 
11 scenes used, i.e.\ scenes 22, 26, 28, 30, 31, 43, 58, 59 70, 75, 77
(all with dialogue, and only dialogue, between Ilsa and Rick).  Rather than 
projections on factors, here the correlations (or cosines of angles with 
factors) are used to directly capture orientation.} 
\label{figcchc}
\end{figure}

In Figure \ref{figcchc}, we look at relatively large 
caesuras.  Consider the partition with five clusters. A horizontal cut of the
dendrogram at level 0.95 provides five clusters.  The five cluster labels are, 
for the 11 scenes: 1, 1, 1, 2, 3, 3, 3, 3, 4, 5, 5.  The fourth and ninth scenes,
viz.\ scene 30 and scene 70, are singletons in this partition.  
There is  discontinuity in 
the evolution of the narrative expressed by these scenes.  

From Figure \ref{figsentiment}, the highlighted boxes refer to the cases of 
scenes 30 and 70.  In Figure \ref{figsentiment} these two scenes are a drifting 
apart of Rick and Ilsa.  Such drifting apart is manifested 
there through somewhat greater distance of the semantic location of the 
terms ``darling'' or ``love'' from the overall semantic location of the 
scene.  These relatively forceful discontinuities, given by the singleton
clusters in the five-cluster partition in Figure \ref{figcchc}, are indicative 
of the changes over scenes 28, 30, 31, and over scenes 59, 70, 75.  

If we look for the next discontinuity, determined by a partition with six clusters, 
and the next partition again with seven clusters, we find the crucial scene to be,
respectively, 26 (the second in sequence) and 31 (the fifth in sequence).  Figure
\ref{figsentiment} provides an insightful perspective on what is happening to 
provide an explanation of these discontinuities. 

In moving from scene 28 to 30, the former here ends with a passionate kiss, 
and the latter is where Ilsa expresses her concern that Rick should 
flee Paris for his safety.  
In moving from scene 30 to 31, and then from 31 to 43, we have scene 31 
where Ilsa and Rick are emotionally close, ending on Ilsa asking Rick to kiss her.

In moving from scene 59 to 70, the former ends in emotionally proximity, 
while scene 70 is where Ilsa expresses her concern that Victor Laszlo needs
to flee Casablanca for his safety.
Then in moving from scene 70 to 75, the latter details the start of the 
departure by plane of Ilsa and Victor.

In the summary of the two parts of Figure \ref{figcchc} that are most at 
issue in our discussion (namely, the transition between scenes 28, 30, 31, 
and the transition between scenes 59, 70, 75), we note affection, and 
non-affection is anxiety or apprehension. 
In line with other work reviewed in section \ref{litreview}, we can easily
 label by scene the degree of affection, including emotional
attachment, of the pair.  This can be accomplished directly from 
Figure \ref{figsentiment}.

\section{Narrative of Emotion in Flaubert's Madame Bovary}
\label{mmebov}

The Casablanca work (i) analyzed strong emotional content, (i) it was based on dialogue,
(iii) while it involved three characters (Ilsa, Victor, Rick), the emotional
narrative was a narrative of attachment of two of these characters counterposed 
to anxiety or apprehension.  In this Flaubert work we use the same methods (i) to 
analyze strong emotional content, (ii) based on dialogue and on extensive description
provided by the novel, and (iii) we have a three-way interplay of the characters
(Emma, Charles and Rodolphe).  Because of the greater amount of data used, 
encompassing the dialogue, the emotional ``signal'' coming from this 
data is faint. We demonstrate the broad and general applicability of our methodology
even with such faint emotional information.  

\subsection{Novel Text Data Used}

In particular because it was used in modelling mood in an interactive game
context, \citet{cavazza09}, we chose to look at the very emotional interplay 
between Emma Bovary and Rodolphe Boulanger in Chapters 9, 10, 11 and 12 of Flaubert's 
19th century novel.  With 
the four selected chapters of {\em Madame Bovary}, we have an initial task 
to address in regard to episodization or segmentation.  \citet{chafe} 
in relating and establishing mappings between memory and story, or narrative, 
considered the following units: (i) memory expressed as a ``disjointed chunk'' of 
text; (ii) episode, expressed by a paragraph; (iii) thought, expressed by a sentence;
and (iv) a focus, expressed by a phrase.  For full generality of implementation, 
we adopted the following approach.  
Taking the four {\em Madame Bovary} chapters, the following preprocessing was
carried out: upper case was set to lower case; punctuation was deleted;
adjacent characters of length 1 or greater were words; and that furnished
3069 unique words, out of 14793 words used in total.  We chose text units of 
20 successive lines, giving rise to 22 successive text segments.  
This input data is available at address www.narrativization.com.

To have sufficient commonality of words, we required 5 occurrences of a word in a 
text segment, and furthermore at least a combined frequency of 5 presences of the word
in all text segments.  That resulted in the 22 text segments being cross-tabulated with 
358 words.  
As previously, using frequency of occurrence values for the 22 text segments, crossed 
by the 358 words, we embedded word profiles, and text segment profiles
in a Euclidean, factor, latent semantic space.  
(The term ``profile'' means the the frequency of occurrence vector, 
divided by the row or column total.  The text segments comprise the rows, and 
the words comprise the columns.) 
From the full-dimensionality factor space, a chronological, sequence-respecting, 
hierarchical clustering was built.  High contributing text segments were determined,
that were more than 3 times the mean text segment contribution.  High contributing words
were similarly determined that were more than 6 times (due to the long tailed 
word frequency distribution) the mean word contribution.  

\subsection{Mapping out the Chronology of the Emotional Triangle}
\label{emottri}

\begin{figure*}
\centering
\includegraphics[width=9cm]{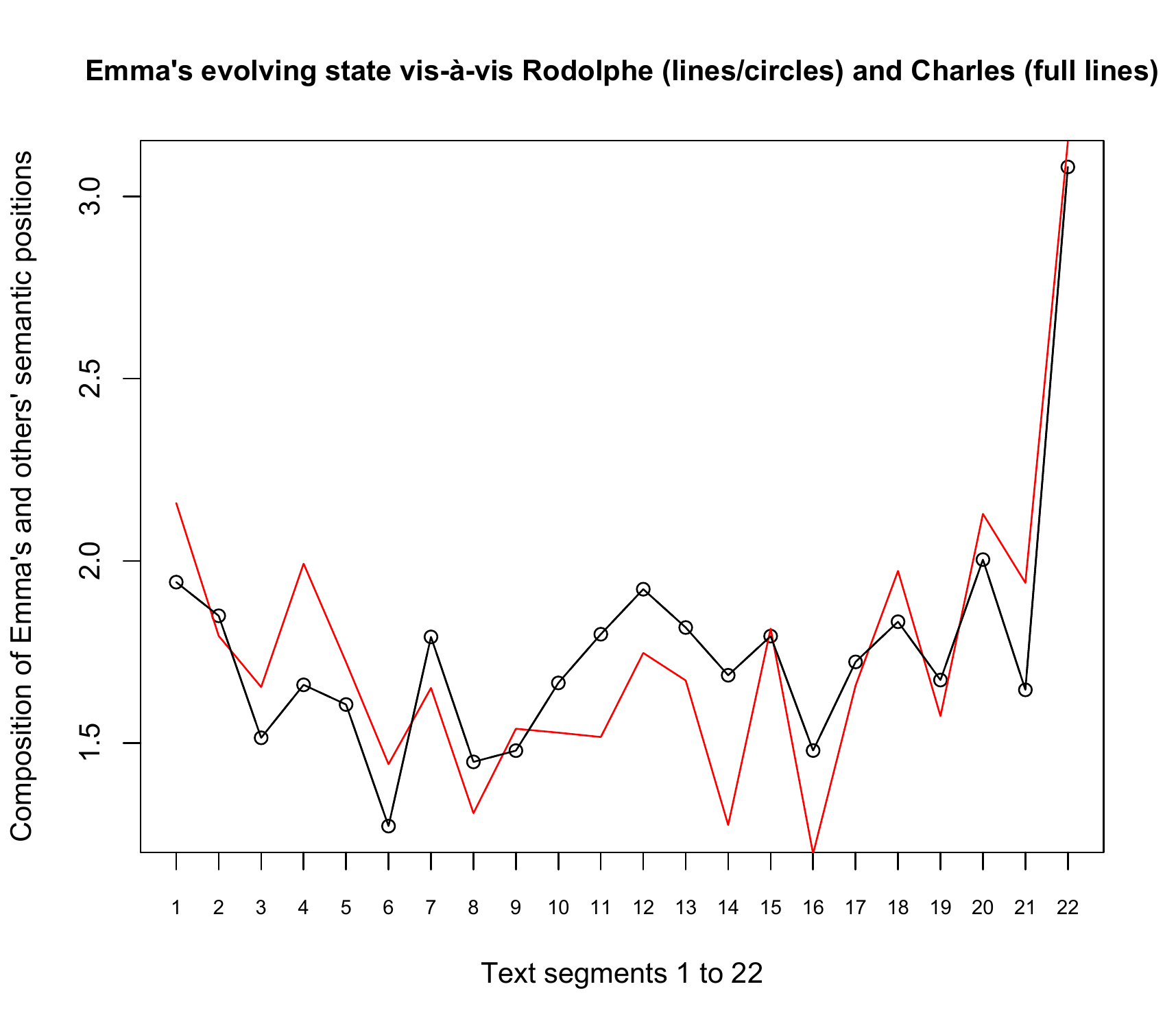}
\caption{The relationship of Emma to Rodolphe (lines/circles, black) and to 
Charles (full line, red) are mapped out. The text segments encapsulate narrative 
chronology, that maps approximately into a time axis.  Low or small values can be viewed
as emotional attachment.}
\label{emroch}
\end{figure*}

Figure \ref{emroch} takes semantic location in the full dimensional 
factor space, and uses the text segment location in this same space as a 
reference point.  

In Figure \ref{emroch}, with three personages, the following was used.  
We determine the distance squared to the text segment's semantic centre of 
gravity.  The distance squared is calculated from the composite vector of Emma 
and Charles, her husband; and the composite vector of Emma and Rodolphe, her
lover.  

Mathematically, and hence algorithmically, this is as follows.  
Let $s_k$ be the $k$th text segment, $f$ is the semantic vector of Emma, 
$m_1$ is the semantic vector of Rodolphe, and $m_2$ is the semantic vector of Charles.
All these vectors are in the semantic or factor space.  Figure \ref{emroch}
shows $ \sqrt{ d^2(s_k, f) + d^2(s_k, m_1) } $ and 
$ \sqrt{ d^2(s_k, f) + d^2(s_k, m_2) } $.  Hence we have composition of the vectors
of Emma and Rodolphe, and of Emma and Charles, in both cases referenced to, or 
centred in, the text segment.   

Figure \ref{emroch} presents an interesting perspective that can be considered
relative to the original text.  Rodolphe is emotionally scoring over 
Charles in 
text segment 1, then again in 3, 4, 5, 6.  In text segment 7, Emma is accosted
by Captain Binet, giving her qualms of conscience.  Charles regains emotional 
ground with Emma through Emma's father's letter in text segment 10, and Emma's
attachment to her daughter, Berthe.  Initially the surgery on Hippolyte in 
text segment 11 draws Emma close to Charles.  By text segment 14 Emma is 
walking out on Charles following the botched surgery.   Emma has total disdain 
for Charles in text segment 15.  In text segment 16 Emma is buying gifts for 
Rodolphe in spite of potentially making Charles indebted.  In text segments 17 
and 18, Charles' mother is there, with a difficult mother-in-law relationship 
for Emma.  Plans for running away ensue, with pangs of conscience for Emma, 
and in the final text segment there is Rodolphe refusing to himself to leave
with Emma.  

We recall that the text segments are somewhat crudely defined by us on the grounds
of being easily defined.  We find nonetheless that they provide interpretable and 
useful narrative flow.  While Figure \ref{emroch} provides a visualization of the 
interplay of the three characters, in Figure \ref{MBsent} we will visualize the
relative strength of emotion.  

\subsection{Tracking the Chronology of Emotional Sentiment}

\begin{figure*}
\centering
\includegraphics[width=9cm]{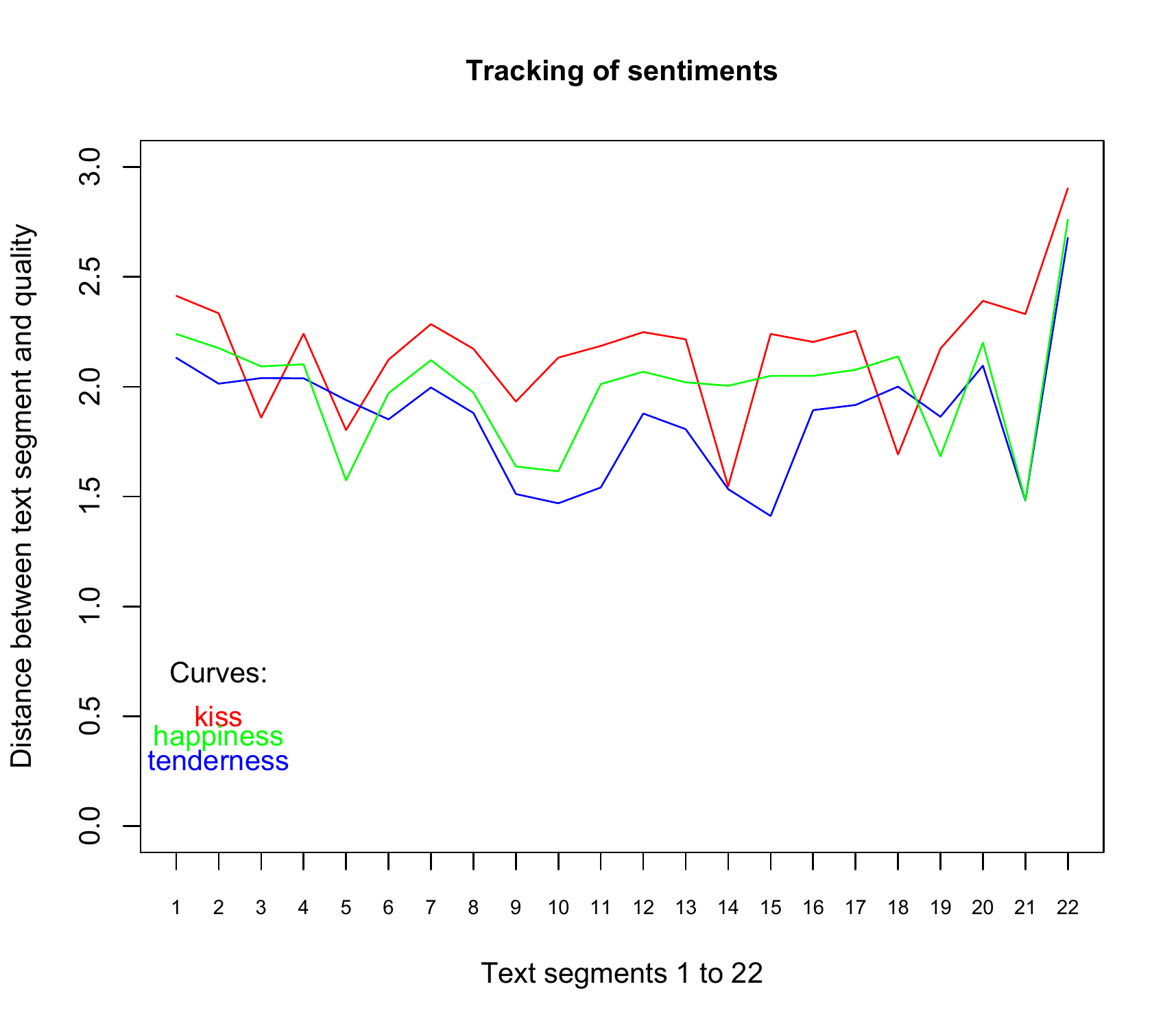}
\caption{A low value of the emotion,
 expressed by the words ``kiss'', ``happiness'' and 
``tenderness'', implies small distance to the text segment. 
These curves, ``kiss'', ``happiness'' and ``tenderness''  start on the upper left
on the top, the middle, and the bottom, respectively.
The chronology of sentiment tracks the closeness of these different 
sentimental terms relative to the narrative, represented by the text segment.  Terms 
and text segments are vectors in the semantic, factorial space, and the full 
dimensionality of this space is used.}
\label{MBsent}
\end{figure*}

Figure \ref{MBsent} displays the evolution of emotional sentiment, expressed by (or 
proxied by) the terms ``kiss'', ``tenderness'', and ``happiness''.   We see 
that some text segments are more expressive of emotion than are other text 
segments.

\section{A General Application: Emotive Terms in Twitter Narratives}
\label{twitter}

Following on from the two case studies that have been detailed, we address the
following.  Can our approach scale up in quantity of data? Can it be applied where
there is no dominant emotional content?  So we want to use our unsupervised 
narrative mapping approach in a general way.  Our motivation is to consider 
data that mixes emotional and non-emotional content.  In this section, we will
be examining a word set of, in all, 22639 words.  

Using the methodology that has been prototyped so far in this article, we now apply
it, at scale, to social media.  To demonstrate no limitations in regard to language,
we use Twitter streams in English, French and German.  To achieve our main objective
of validating our methodology, news-related Twitter sources were taken.  While very
minimal in regard to emotional content as such, this data provides us with a useful
validation framework.  

Validation here is to show how emotive terms (i) are baselined against the narrative
flow, and (ii) can be contrasted, one to another, against this baselining.

We took the most recent 1000 tweets from the Twitter sources of the newspapers:
New York Times, {\tt @nytimes}; Le Monde, {\tt @lemondefr} (in French);
Guardian (UK), {\tt @guardian};
Irish Times (Ireland), {\tt @IrishTimes}; S\"uddeutsche Zeitung, {\tt @SZ} (in German).
We captured
these tweet streams on 2014-10-17 (between 19:02 and 19:38 UTC).  We have made
this data available
at www.narrativization.com.  The 1000 tweets went back in time to between 2014-10-08 and
2014-10-13 (for, respectively, the Irish Times and the Guardian; this shows a slightly
greater degree of tweeting activity in the latter case).

For processing and matching convenience, we replaced the character ``@'' signifying a
tweeter name by ``XYZ'', and the character ``\#'' signifying a topic by ``ZYX''.  HTML
ampersand was replaced by the word for ``and'' in the appropriate language.  The
apostrophe was replaced by a blank.  All punctuation, and numeric characters, were
deleted.  All upper case was set to lower case.  Stopwords, in the relevant language,
and available from the SMART information retrieval system and accessed using the R
{\tt tm}, Text Mining Package, were removed.  In Table \ref{newspapers} the number of
words thus determined are listed for the 1000 selected tweets.  Mostly,
the frequency of occurrence of a word in a tweet was just 1.  The density of non-zero
values in the tweets $\times$ words data is reported on in Table \ref{newspapers}.

Since our analytics are applied on the tweet stream, it is important that there be some
commonality of words, rather than, say, a once and once only use of a word.  From the
experience of analyzing many types of text including Twitter, required thresholds are used
of (i) word use in at least five tweets, and (ii) at least five uses of the word.  In
Table \ref{newspapers} there is note of the words retained that pass these thresholds.
Thus the long tail of the
word frequency distribution is cut.  A consequence of this is that tweets may have all
their (low frequency of occurrence) words removed.  So Table \ref{newspapers} indicates
the number of (now non-empty) tweets that were retained.

A metric embedding of the retained tweet cloud, and of the retained word cloud, was
determined using Correspondence Analysis.  Using the full dimensionality, in the
Correspondence Analysis factor space endowed with the Euclidean distance, the
retained tweet stream is segmented.  For this, the complete link hierarchical
agglomerative hierarchical clustering is used, with the following modification,
as described in section \ref{method}. 
Rather than pairwise agglomerations until all
the (contiguous) tweet clusters are in one overall cluster, here we statistically assess
whether an agglomeration is to be permitted.  Permitting an agglomeration is based on a
permutation test of pairwise distances from randomization of groups.
A 10\% significance level is used.  This results in a partition of the tweet sequence.
We refer to these tweet clusters as segments in the stream of tweets.  The numbers of
segments found in the tweet streams are noted in Table \ref{newspapers}.
Singleton segments, or segments with a small number of tweets, may be of interest in
terms of exceptionality but, here, our objective is to summarize the tweet flow by
means of the successive tweet segments. For this reason, we restrict our attention
to tweet segments that have more than one tweet.  Doing this implies again that some
words retained are no longer present.  (That a segment has just one tweet is most
consistent with relatively rare words being in use by that segment.)

These tweet flow segments define our Twitter narrative for us. They provide a particular
summarization of the Twitter narrative.  In a very high level sense, they respond to our
need to periodize or break into episodes the narrative flow.  (Indeed, Aristotle's
{\em Poetics}, dating from c.\ 350 BC, refers to episodization, whereby story or
narrative is developed through episodes that are elaborated on.)

\begin{table*}
\begin{center}
\begin{tabular}{lcccccc} \hline                                                             
Source            &     Tweets & Words &  Density  &       Tweets(2) & Words(2) &          
 Segments  \\ \hline                                                                       
NY Times    &     1000  & 4153   & 0.00183   &      846   &    247, 247  &  31 \\    
Le Monde          &     1000  & 4428   & 0.00176   &      852   &    245, 234  &  43 \\    
Guardian          &     1000  & 4637   & 0.00180   &      907   &    267, 267  &  51 \\
Irish Times       &     1000  & 3870   & 0.00207   &      917   &    275, 269  &  50 \\
S\"udd. Z. &  1000 &  5551  & 0.00163   &      836   &    240, 230  &  49 \\
\hline
\end{tabular}
\end{center}
\caption{Summary of data properties.  Tweets: initial number of tweets.  Words:
total number of words.  Density: of tweets $\times$ words data.  Tweets(2): non-empty
tweets, given sufficient word occurrences.  Words(2): number of sufficiently
occurring words retained, followed by number
of words for non-singleton tweet segments.  Segments: tweet segments in tweet stream.}
\label{newspapers}
\end{table*}

\begin{table*}
\begin{center}
\begin{tabular}{llccc} \hline
Tweet    &    Emotive  &  Tweet Segment      &   Tweet Segment        &    Tweets in  \\
Source   &    Term &      Start              &   Finish               &    Segment \\
\hline
NY Times &  hysteria &    2014-10-09 07:01:26 &  2014-10-09 09:49:21  &      13 \\
NY Times &  threats  &    2014-10-15 16:59:11 &  2014-10-15 21:00:18  &      37 \\
Le Monde &   col\`ere &   2014-10-15 16:58:23 &  2014-10-15 16:58:23  &      17 \\
Le Monde &   espoirs  &   2014-10-12 03:35:10 &  2014-10-13 08:13:53  &      36 \\
Guardian &   worst    &   2014-10-14 09:55:23 &  2014-10-14 12:28:10  &      60 \\
Guardian &   threat   &   2014-10-15 18:15:32 &  2014-10-16 05:35:02   &     10 \\
Irish Times & concerns &   2014-10-15 06:04:05 &  2014-10-15 12:06:23  &      42 \\
Irish Times & crisis  &    2014-10-14 05:41:34 &  2014-10-14 07:17:18  &      16 \\
S\"udd. Z. &  beste   &   2014-10-08 15:45:13  & 2014-10-08 18:24:14   &     10 \\
S\"udd. Z. &  erschreckende &  2014-10-04 06:37:53 & 2014-10-04 14:16:59 &   16 \\
\hline
\end{tabular}
\end{center}
\caption{Emotive words (``col\`ere'' = anger; ``espoirs'' = hopes; ``beste'' = best;
``erschreckende'' = frightful).  The Twitter narrative flow has been decomposed
into narrative segments, see Table \ref{newspapers}.  The closest segment to the
emotive word is reported here.  Semantic distance is used between the word and the
tweet segment.}
\label{emotiveword}
\end{table*}

Now we select emotive terms (notwithstanding the emotionally poor context, although
this is helpful in limiting our choice of emotive terms).  
The semantically closest tweet segment to the term is noted.  The tweet segment is
noted using the time of the initial tweet, and the time of the last tweet.  Universal
time is used, UTC. The number of tweets in that semantically closest tweet segment are
noted too.  Our results are reported on in Table \ref{emotiveword}.

\begin{figure}
\centering
\includegraphics[width=8cm]{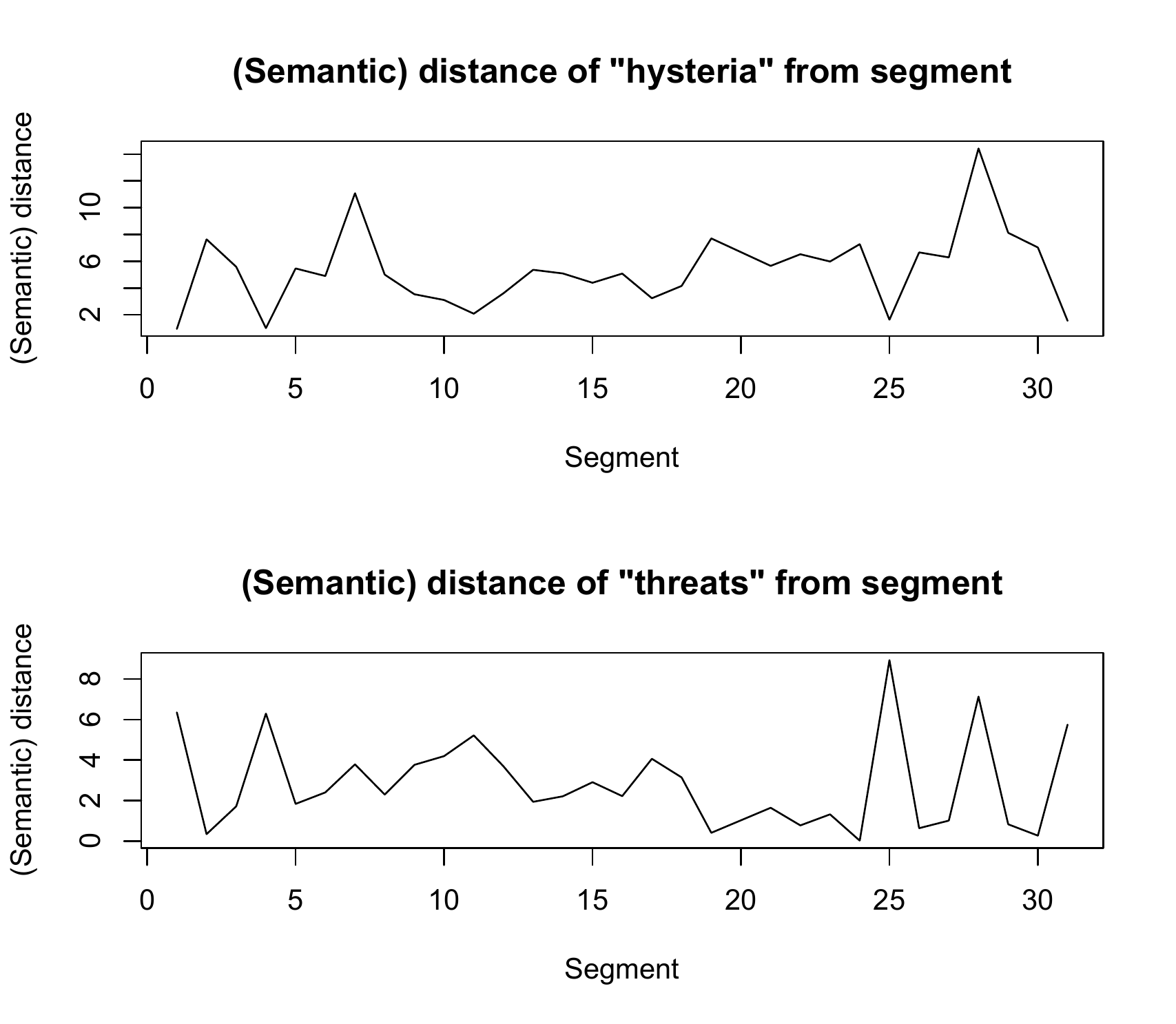}
\caption{From 1000-tweet stream of the New York Times, {\tt @nytimes}.
The Euclidean distances between the word and tweet segments 1 to 31.
All words and all tweet segments are embedded in the Correspondence Analysis factor
space.  This latent semantic space is endowed with the Euclidean metric and all
points (words, tweet segments) are of equal mass.  To aid in visualization, the
distances are connected in a continuous curve.  Small values of distance indicate
a close semantic association of word and tweet segment.}
\label{figNYT}
\end{figure}

In all cases, the semantic distance to the tweet segment (using its semantic location)
was plotted for the selected emotive terms.  Figure \ref{figNYT} shows the outcome for
one of the newspaper Twitter sources.  Our emotive terms were selected by us for their
positive or negative tendency, and it can be noted how this contrast is found in
 Figure \ref{figNYT}.  The (emotively) more positive leaning words have quite
contrasting curves, i.e.\ distances to tweet segments, than is the case for the
more negative leaning words.

As we have shown, we locate semantically the selected terms, here emotive terms,
in the narrative framework.  The narrative framework is segmented, or if desired,
hierarchically structured.  Emotive terms like those used here are indicators of
what we seek to study.  In earlier sections, indicators were of amorous or other
personal attachment.  In the present section, dealing with news of national and
international portent, indicators were of happier (or relaxed) versus sadder 
(or disturbing) periods.  Finally, in this article, the
narrative used has been literary story-telling, or cinematic script, or information
dissemination.

Given the relevant data, further application of our work could be to the
narrative of emotion, and action, for individuals or for groups of individuals.

\section{Conclusions}

Innovation in this work includes the determining of subplots of the overall narrative
in section \ref{subplot}.  In section \ref{sectVC} we develop further our approach
for multiple scale, chronological representation of the narrative.  In section 
\ref{emottri} we have a new approach that displays the chronology of a
3-way emotional relationship.  In section \ref{twitter} we open up our innovative 
methodology to the analysis of narrative in any codified form, expressed as text 
or otherwise.  

We have exemplified our data driven approach to semantically mapping out narrative 
flow.  By ``semantics'' we mean the entirety of given context.  Any given thing 
is at the focal point of the totality of its interrelationships with all other 
things.  Our mathematical and computational methodology allows us to structure 
narrative.  

We applied this to (i) human dialogue, (ii) descriptive story, and (iii) 
streaming of information.   

By semantically mapping out and structuring the narrative flow, we were able to 
semantically locate emotionally laden terms, or potentially emotive terms. 

Potential applications of this work, that were noted in section \ref{litreview},
include gamification, and social behaviour analysis and forensics.  Other potential
applications that we may note are new vistas in computational humanities, and 
computational psychoanalysis (cf.\ \citet{murtagh2014}).  

All data used in this article are available online, in the interest of reproducibility 
of results and of our findings.  We have demonstrated the following in this work on 
literature and social media case studies.  Our methodological high points are: 
(i) the potential of semantic mapping for theme tracing, and (ii) for theme subdivision; 
and also (iii) content-based segmenting of the flow of narrative.  

\bibliographystyle{plainnat}
\bibliography{murtagh_ganz_new_v20}

\end{document}